\documentclass[conference]{IEEEtran}

\usepackage{cite}
\usepackage{amsmath, amssymb, amsfonts}
\usepackage{graphicx}
\usepackage{textcomp}
\usepackage{booktabs}
\usepackage{threeparttable}
\usepackage{tabularx}
\usepackage{xcolor}
\usepackage{multirow}
\usepackage{soul}
\usepackage{times}
\usepackage{latexsym}
\usepackage[T1]{fontenc}
\usepackage[utf8]{inputenc}
\usepackage{microtype}
\usepackage{enumitem}
\usepackage{inconsolata}
\usepackage{pifont}
\usepackage{float}
\usepackage{afterpage}
\usepackage{placeins}
\usepackage{adjustbox}
\usepackage{newunicodechar}
\newunicodechar{≥}{$\geq$}
\usepackage[framemethod=TikZ]{mdframed}
\usepackage{multicol}
\usepackage[most]{tcolorbox}
\usepackage[linesnumbered,ruled,vlined]{algorithm2e} 
\usepackage{hyperref}
\IEEEoverridecommandlockouts

\newcommand\copyrighttext{%
  \footnotesize \textcopyright 2025 IEEE. Personal use of this material is permitted.
  Permission from IEEE must be obtained for all other uses, in any current or future
  media, including reprinting/republishing this material for advertising or promotional
  purposes, creating new collective works, for resale or redistribution to servers or
  lists, or reuse of any copyrighted component of this work in other works.
  DOI: 10.1109/CAI64502.2025.00080}
\newcommand\copyrightnotice{%
\begin{tikzpicture}[remember picture,overlay]
\node[anchor=south,yshift=10pt] at (current page.south) {\fbox{\parbox{\dimexpr\textwidth-\fboxsep-\fboxrule\relax}{\copyrighttext}}};
\end{tikzpicture}%
}

\def\BibTeX{{\rm B\kern-.05em{\sc i\kern-.025em b}\kern-.08em
    T\kern-.1667em\lower.7ex\hbox{E}\kern-.125emX}}
\begin{document}

\title{A Layered Multi-Expert Framework for Long-Context Mental Health Assessments\\}

\author{
    \IEEEauthorblockN{Jinwen Tang\IEEEauthorrefmark{1}, Qiming Guo\IEEEauthorrefmark{2}, Wenbo Sun\IEEEauthorrefmark{3}, and Yi Shang\IEEEauthorrefmark{1}}
    
    \IEEEauthorblockA{\IEEEauthorrefmark{1}\textit{Electrical Engineering and Computer Science Department} \\
    \textit{University of Missouri}, Columbia, Missouri, USA \\
    \{jt4cc, shangy\}@umsystem.edu}

    \IEEEauthorblockA{\IEEEauthorrefmark{2}\textit{Department of Computer Science} \\
    \textit{Texas A\&M University-Corpus Christi}, Corpus Christi, Texas, USA \\
    qguo2@islander.tamucc.edu}

    \IEEEauthorblockA{\IEEEauthorrefmark{3}\textit{Department of Computer Science} \\
    \textit{Delft University of Technology}, Delft, Netherlands \\
    w.sun-2@tudelft.nl}
    \thanks{Note: The first two authors contributed equally to this work.}
}

\maketitle

\copyrightnotice

\begin{abstract}
Long-form mental health assessments pose unique challenges for large language models (LLMs), which often exhibit hallucinations or inconsistent reasoning when handling extended, domain-specific contexts. We introduce \textit{Stacked Multi-Model Reasoning (SMMR)}, a layered framework that leverages multiple LLMs and specialized smaller models as coequal "experts". Early layers isolate short, discrete subtasks, while later layers integrate and refine these partial outputs through more advanced long-context models. We evaluate SMMR on the DAIC-WOZ depression-screening dataset and 48 curated case studies with psychiatric diagnoses, demonstrating consistent improvements over single-model baselines in terms of accuracy, F1-score, and PHQ-8 error reduction. By harnessing diverse "second opinions", SMMR mitigates hallucinations, captures subtle clinical nuances, and enhances reliability in high-stakes mental health assessments. Our findings underscore the value of multi-expert frameworks for more trustworthy AI-driven screening.
\end{abstract}

\begin{IEEEkeywords}
NLP, LLM, Artificial Intelligence, Mental Health, Psychiatry, Explainable AI, Multi-Model Reasoning
\end{IEEEkeywords}

\section{Introduction}

Mental health remains a pressing public health concern, complicated by stigma, limited clinical services, and socioeconomic barriers \cite{gulliver2010perceived, pescosolido1999people}. Advances in large language models (LLMs) such as GPT and Mistreal offer potential for discreet, scalable mental health screenings \cite{openai2023gpt}, helping individuals who might otherwise be reluctant to seek professional help due to judgment or resource constraints.

However, using LLMs in high-stakes mental health assessments poses distinct challenges. First, \textbf{mental health data are inherently subjective}: while instruments like Patient Health Questionnaire–8 (PHQ-8) provide structured guidelines, the interpretation of symptoms can vary significantly from one conversation or individual to another \cite{ng2019provider, Marsolek20240969PD}. Subtle linguistic nuances and personal expressions often shape how distress is perceived. Second, \textbf{long and complex transcripts}—such as multi-turn interviews or narrative case studies—can overwhelm advanced models, leading to hallucinations or inconsistencies \cite{huang2023survey, dahl2024large}. Third, \textbf{specialized tools} like PHQ-8 offer only partial mitigation; a single LLM “expert” might still fail to capture nuanced details without additional context or repeated checks \cite{tangshang2024advancing}. Although iterative reassessments and third-party evaluations can refine accuracy, they often rely on structured prompts or assume a known “best” configuration.

In response, we propose \textit{Stacked Multi-Model Reasoning (SMMR)}, a framework in which diverse “expert” models collectively assess the same input without any being deemed categorically superior. SMMR leverages multiple reasoning steps, consolidating “second opinions” in a layered manner:
\begin{enumerate}[noitemsep, topsep=0pt] 
\item \textbf{Multiple Experts, One Pipeline:} Early layers employ smaller or specialized LLMs to generate preliminary assessments, treating each as an independent perspective.
\item \textbf{Iterative Refinement:} Subsequent layers integrate and reconcile these varying outputs, using long-context models to produce a cohesive final judgment.
\item \textbf{Robust Mental Health Screening:} By adopting a multi-expert approach, SMMR reduces hallucinations and inconsistencies common in extended conversations, improving diagnostic reliability for clinical interviews and narrative case studies.
\end{enumerate}

We evaluate SMMR on two complex mental health datasets: the DAIC-WOZ dataset \cite{gratch2014distress}, designed for studying psychological distress (anxiety, depression, PTSD), and a curated set of narrative case studies drawn from real-world clinical scenarios. Our experiments demonstrate that SMMR offers a practical way to provide “second opinions” within a unified workflow—\textit{without} requiring a single preselected model or intricate prompt engineering \cite{xu2024mental, yu2024experimental}. By treating all models initially as independent experts, then effectively aggregating imperfect prompts and diverse outputs, SMMR navigates the subjective and nuanced landscape of mental health assessments more reliably, pointing toward safer AI-assisted screenings.

\section{Background}

Artificial intelligence (AI) is reshaping mental health care by providing increasingly sophisticated analytical tools and treatment supports. Early systems relied on simple word-counting and text analysis \cite{tausczik2010psychological}, but the advent of large language models (LLMs) has radically expanded these capabilities. Modern AI-driven applications—such as semi-automated screening platforms, interactive self-reporting interfaces, and responsive mental health chatbots—now enable more nuanced interpretation of human language and broader access to mental health services \cite{li2023systematic}.

Despite these advances, many solutions remain constrained by short-context analyses or single-model pipelines. For instance, Ohse et al.\ \cite{ohse2024zero} investigate NLP models like BERT and GPT-4 for depression detection, yet their approach primarily relies on individual model outputs. Agrawal \cite{agrawal2024illuminate} highlights the potential of enhanced prompt engineering to improve explainability and intervention planning in mental health AI, though it too centers on a single-model framework. Similarly, Tang and Shang \cite{tangshang2024advancing} propose a GPT-based system for pre-screening mental health disorders, demonstrating that domain-specific fine-tuning can improve early detection. However, these methods do not fully address the extended, subjective nature of clinical transcripts or case studies where long-format reasoning and multiple perspectives may be crucial.

In the following section, we introduce \textit{Stacked Multi-Model Reasoning (SMMR)}, a layered framework designed to overcome the limitations of single-model or short-context approaches. By integrating multiple “expert” LLMs and specialized models within a unified pipeline, SMMR offers more robust error checks, interpretive consistency, and the capacity to reconcile diverse viewpoints in subjective mental health data.

\section{Method}
\label{sec:method}

\subsection{Stacked Multi-Model Reasoning (SMMR)}
\label{subsec:SMMR}

\begin{figure}[ht]
    \centering
    \includegraphics[width=0.4\textwidth]{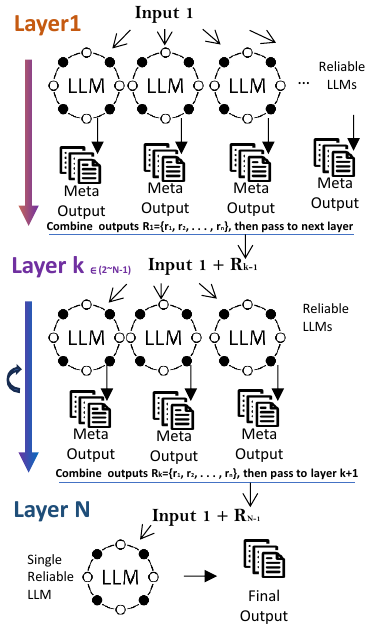} 
    \caption{Overview of the Stacked Multi-Model Reasoning (SMMR) framework. Each layer refines or integrates outputs from the previous layer to form the next layer’s input. No prior ranking of model quality is assumed.}
    \label{fig:smmr_framework}
\end{figure}

\textbf{Motivation and Overview:}
Long-form mental health assessments often require careful synthesis of complex, multi-turn data. Even advanced Large Language Models (LLMs) can become prone to hallucinations or inconsistent reasoning in these extended contexts. To address this problem, we propose \textit{Stacked Multi-Model Reasoning (SMMR)}, a framework that treats each LLM (or smaller specialized model) as an independent ``expert,'' without \textit{a priori} knowledge of which model might be ``best.'' By layering multiple models---each offering a second opinion---SMMR leverages collective insights to mitigate the weaknesses of any single model. Figure~\ref{fig:smmr_framework} illustrates the conceptual architecture of SMMR, while Algorithm~\ref{alg:SMMR} provides pseudocode for the overall process.

\begin{algorithm}[ht]
\caption{Stacked Multi-Model Reasoning (SMMR)}
\label{alg:SMMR}
\SetAlgoLined
\KwIn{Long-context data $\mathcal{X}$; set of $L_1$-type Single-Step Models $\{M_1^1,\dots,M_1^n\}$; set of $L_k$-type Long-Context Models $\{M_k^1,\dots,M_k^m\}$ for each layer $k$; number of layers $N$.}
\KwOut{Final aggregated output $R_{\mathrm{final}}$.}

\textbf{Layer 1: Initial Opinions} \\
\ForEach{model $M_1^i \in \{\text{Single-Step Models}\}$}{
    $r_1^i \gets M_1^i(\mathcal{X})$\;
}
Aggregate outputs: $R_{1} \gets \{r_1^1, r_1^2,\dots,r_1^n\}$\;

\textbf{Layers 2 to $N-1$: Iterative Refinement} \\
\For{$k = 2$ to $N - 1$}{
  \ForEach{model $M_k^j \in \{\text{Long-Context Models at layer }k\}$}{
    $r_k^j \gets M_k^j(R_{k-1})$\; 
  }
  Aggregate outputs: $R_k \gets \{r_k^1, r_k^2,\dots,r_k^m\}$\;
}

\textbf{Layer N: Final Consolidation} \\
Use a single, reliable Long-Context Model $M_N^*$: \\
$R_{\mathrm{final}} \gets M_N^*(R_{N-1})$\;

\Return $R_{\mathrm{final}}$ 
\end{algorithm}

\paragraph{Layer 1: Multiple Independent Experts.} 
In this initial step, SMMR applies multiple \emph{single-step} models, each independently processing the input $\mathcal{X}$. These models could be smaller LLMs or specialized classifiers that excel at short-context tasks. Since we do not assume any model to be inherently superior, all outputs are aggregated on equal footing, forming $R_1$ for further refinement.

\paragraph{Layers 2 to $N-1$: Iterative Refinement with Long-Context Models.}
Subsequent layers introduce \emph{long-context} LLMs capable of handling extended or detailed inputs. Each model in layer $k$ takes the aggregated output $R_{k-1}$ from the previous layer and refines it, generating new outputs $r_k^j$. These outputs are then aggregated into $R_k$. The iterative process allows each layer to provide a ``second opinion,'' reconciling discrepancies and minimizing hallucinations or inconsistencies that could arise from any single model.

\paragraph{Layer N: Final Consolidation by a Reliable Model.}
In the last layer, a single, long-context \textit{Reliable Model} $M_N^*$ synthesizes the refined outputs from $R_{N-1}$ into the ultimate result $R_{\mathrm{final}}$. This top-tier model is selected for its stable performance on nuanced, long-context data—making it especially well-suited for high-stakes mental health evaluations.

\paragraph{Dynamic Stopping Based on Performance Optimization.}
To ensure optimal performance and computational efficiency, the SMMR framework incorporates a dynamic stopping mechanism. After each layer processes and refines the aggregated outputs from the preceding layer, the framework evaluates the current performance using predefined metrics (e.g., Accuracy, F1-score, MAE). If the performance metrics improve compared to the previous layer by a significant threshold $\delta$, SMMR proceeds to the next layer for further refinement. This iterative process continues until adding another layer does not result in performance gains beyond the threshold. The final output used for evaluation is thus the result from the layer that achieved the highest performance metrics, optimizing both accuracy and resource utilization.

\medskip

By dividing the reasoning process into distinct layers and integrating diverse model outputs at each stage, SMMR effectively mitigates the risk of hallucinations and maintains stronger consistency for complex mental health tasks.

\subsection{Datasets and Task Setup}
\label{subsec:datasets_and_tasks}

In order to demonstrate the effectiveness of our Stacked Multi-Model Reasoning (SMMR) framework, we evaluate on two complementary mental health datasets: (1) an externally sourced dataset (DAIC-WOZ) \cite{gratch2014distress}, and (2) a curated collection of narrative case studies. Our primary task involves predicting mental health risk and severity, operationalized through PHQ-8 scores or binary labels, based on extended transcripts or descriptive case data. By adopting a layered approach that draws on multiple ``expert'' models, SMMR aims to reduce hallucinations, enhance diagnostic fidelity, and improve the reliability of long-context LLM-based assessments.

\textbf{The DAIC-WOZ Database:} A private dataset consists of 187 labeled interviews designed to assess psychological distress such as anxiety, depression, and PTSD. Each interview includes a PHQ-8 score and its corresponding binary label. Following the standard protocol in \cite{gratch2014distress}, we split the dataset into training and testing subsets. To create a more realistic, long-context input for our models, each conversation was concatenated into a single data stream by aligning segments according to the speaker’s starting time. The consolidated dataset includes references to the speaker, the content of each segment, and punctuation marks consisting of a period followed by a slash (./) to denote the end of each speaking turn. We evaluate model outputs using:
\begin{itemize}[leftmargin=*]
    \item \textbf{PHQ-8 Score Estimation (0--24)}: We compare the predicted PHQ-8 score to the ground truth, measuring accuracy with metrics such as Mean Absolute Error (MAE) and Root Mean Squared Error (RMSE).
    \item \textbf{Binary Classification (PHQ-8 $\geq 10$)}: The model’s output is thresholded at 10 to indicate the presence or absence of clinically significant depressive symptoms.
\end{itemize}

\textbf{Case Study Dataset:} To complement the interview-style data from the DAIC-WOZ database, we collected 48 narrative case studies featuring professionals' concluded psychiatric diagnoses. These cases, sourced from academic texts and clinical literature, include detailed demographic, behavioral, and contextual information. We manually extracted binary conclusions and types of disorders from the diagnoses provided by clinical professionals to verify the presence or absence of mental health concerns. Of these case studies, six are formatted as conversational transcripts, while the remaining 42 are presented in a descriptive format. An example is shown in Table~\ref{tab:case_study_example}.

Because these cases do not provide PHQ-8 labels, we adopt an alternative evaluation scheme:
\begin{itemize}[leftmargin=*]
    \item \textbf{Mental Concern (0, 1, 2)}: Determines whether the case indicates no mental health issue (0), presence of a mental health issue (1), or if the conclusion is indeterminate from the data (2).
    \item \textbf{Disorder-Type Identification}: Extracts the specific mental health disorder(s) (if any) from the text. We measure accuracy by comparing the model-identified disorders to a reference list of ground-truth labels, considering minor variations in naming as valid matches.
\end{itemize}

\begin{table}[htbp]
\centering
\caption{Case Study Example}
\label{tab:case_study_example}
\begin{tabular}{@{}p{\columnwidth}@{}}
\hline
\textbf{Case:} A 37-year-old white male infantryman stationed in Iraq arrived at a field hospital complaining that his superior officer placed poisonous ants in his helmet. His face is covered with excoriations from persistent scratching. On further examination, he is stuporous and has mildly slurred speech, tremor, and mint odor to his breath. Later his troop leader mentioned that his Humvee was littered with empty bottles of mouthwash and that the man has been reprimanded for falling asleep at his post. After a night of rest, he discussed his excessive use of mouthwash in place of alcohol, which is the only available form of alcohol in Iraq. \\ \hline
\textbf{Conclusion:} The individual in the provided case study exhibits symptoms consistent with delusions of persecution (believing his superior officer placed poisonous ants in his helmet), excoriations from persistent scratching indicative of possible hallucinations or delusions, stupor, slurred speech, tremors, and the smell of mint on his breath. Additionally, his excessive use of mouthwash as a substitute for alcohol suggests a coping mechanism or self-medication behavior. While the text does not explicitly provide a diagnosis, the symptoms described align with features of psychosis and substance abuse, possibly indicating comorbid conditions such as schizophrenia spectrum disorder and alcohol use disorder. \\ \hline
\end{tabular}
\end{table}

\begin{figure*}[!ht] 
  \centering
  \includegraphics[width=0.75\linewidth]{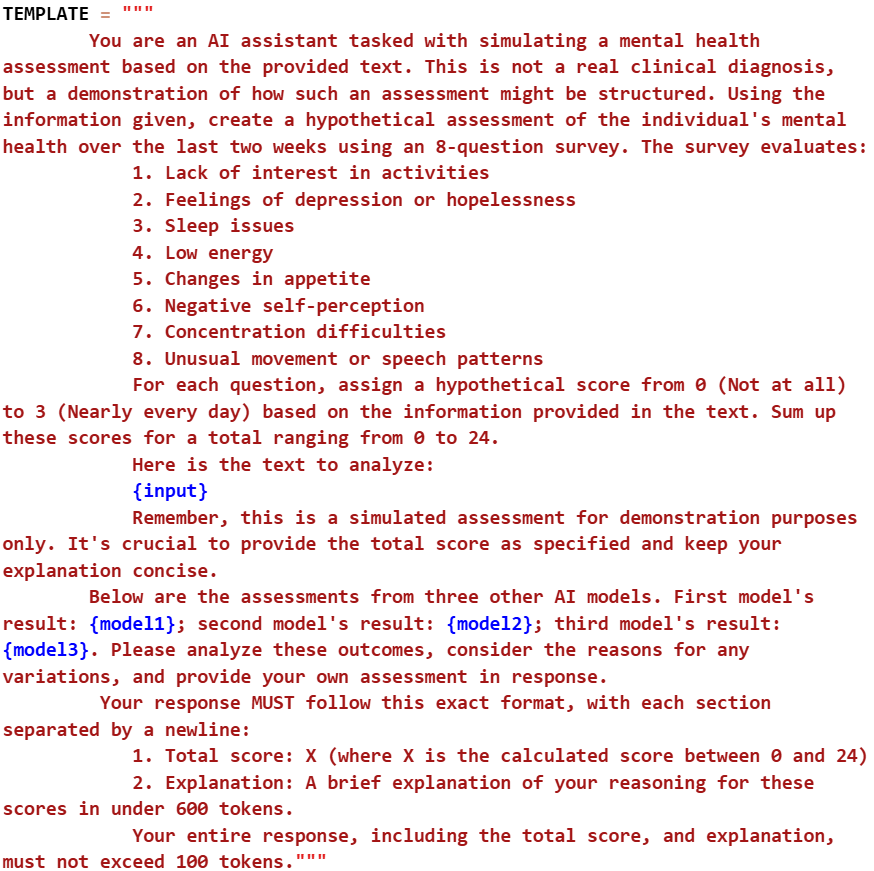}
  \caption{SMMR Prompt for DAIC-WOZ Dataset}
\label{fig:Prompt_for_DAIZ_data_SMMR}
\end{figure*}

\begin{figure*}[!ht] 
  \centering
  \includegraphics[width=0.75\linewidth]{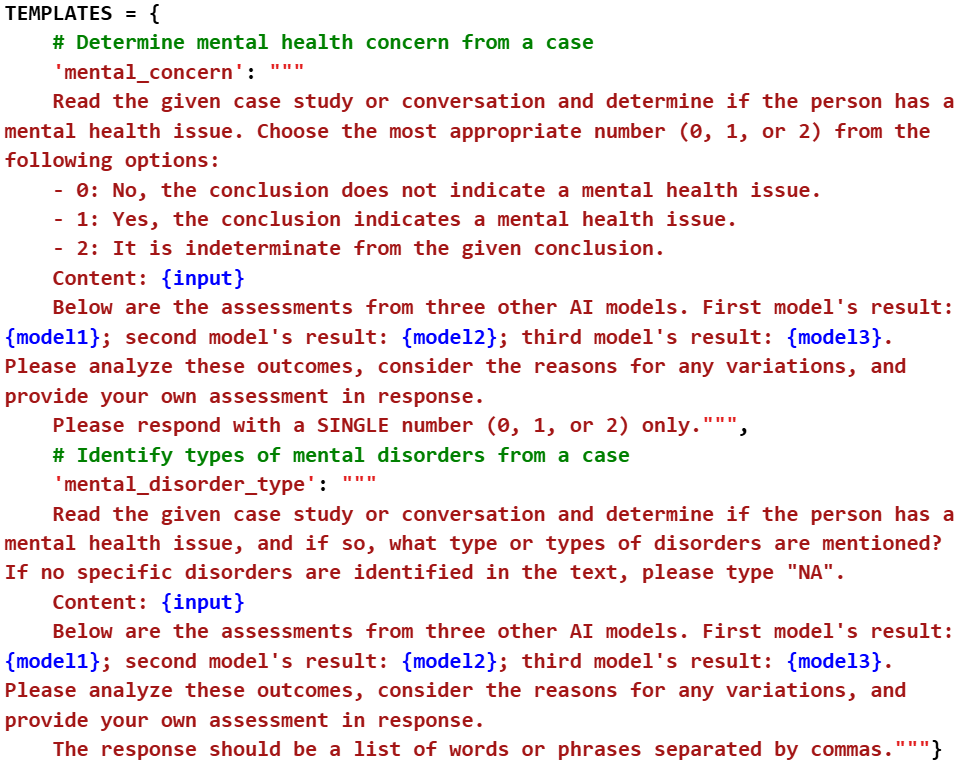}
  \caption{SMMR Prompt for Case Study Dataset}
\label{fig:Figure_Case_Study_SMMR}
\end{figure*}

\textbf{Model Selection:} We initially conducted a pilot test with the smaller, local Mistral model to explore on-premise feasibility. However, the majority of outputs were invalid or incomplete, likely due to context window constraints and the domain-specific nature of the data. Consequently, we omitted local models and go with GPT models but note that future work could revisit them if they become more robust.

\textbf{To Evaluate:} For each dataset instance—whether an interview transcript from DAIC-WOZ or a narrative case study—SMMR processes the \emph{entire} text in a layered manner (the prompts shown in Figure \ref{fig:Prompt_for_DAIZ_data_SMMR} and Figure \ref{fig:Figure_Case_Study_SMMR}, respectively):
\begin{enumerate}[leftmargin=*]
    \item \textit{Layer 1} applies multiple smaller or specialized models in parallel to gather initial assessments.  
    \item \textit{Layers 2 through $N-1$} employ advanced long-context LLMs to refine, reconcile, and aggregate these preliminary outputs.  
    \item \textit{Layer $N$} employs a single, reliable long-context model to finalize the assessment and generate the ultimate decision, whether it be a binary classification or a PHQ-8 score.
    \item \textit{Dynamic Stopping}: The SMMR framework automatically terminates additional layering when further layers do not yield performance improvements beyond a predefined threshold.
\end{enumerate}

This pipeline effectively creates multiple “checkpoints” for error correction and multi-expert verification, ultimately aiming to improve the reliability of mental health evaluations in lengthy and complex conversations.

\begin{table*}[!t]
\centering
\caption{Performance Enhancement for SMMR Across DAIC-WOZ Subsets}
\label{tab:smmr}
\begin{threeparttable}
\fontsize{10.5}{10}\selectfont 
\setlength{\tabcolsep}{1.5pt} 
\begin{tabular}{l|l|lccccccccc}
\toprule
\textbf{Dataset} & \textbf{Model} & \textbf{Method} & \textbf{Acc.} & \textbf{F1} & \textbf{MF1.} & \textbf{MPrec.} & \textbf{MRec.} & \textbf{ROC AUC} & \textbf{MAE} & \textbf{RMSE} \\
\midrule
\multirow{4}{*}{Training} & \multirow{2}{*}{GPT-3.5-turbo} & Baseline & 0.51 & 0.49 & 0.51 & 0.60 & 0.61 & 0.61 & 5.23 & 6.17 \\
                          &                                & SMMR Enhanced & \textbf{0.69} & \textbf{0.59} & \textbf{0.67} & \textbf{0.68} & \textbf{0.72} & \textbf{0.72} & \textbf{4.19} & \textbf{5.18} \\
                          & \multirow{2}{*}{GPT-4-turbo}   & Baseline & 0.76 & 0.64 & 0.73 & 0.72 & 0.75 & 0.75 & 3.32 & 4.07 \\
                          &                                & SMMR Enhanced & \textbf{0.79} & \textbf{0.69} & \textbf{0.76} & \textbf{0.75} & \textbf{0.79} & \textbf{0.79} & 3.33 & 4.24 \\
\midrule
\multirow{4}{*}{Testing}  & \multirow{2}{*}{GPT-3.5-turbo} & Baseline & 0.55 & 0.57 & 0.55 & 0.70 & 0.68 & 0.68 & 6.04 & 6.81 \\
                          &                                & SMMR Enhanced & \textbf{0.76} & \textbf{0.70} & \textbf{0.75} & \textbf{0.76} & \textbf{0.81} & \textbf{0.81} & \textbf{4.22} & \textbf{5.54} \\
                          & \multirow{2}{*}{GPT-4-turbo}   & Baseline & 0.77 & 0.67 & 0.74 & 0.74 & 0.77 & 0.77 & 3.21 & 4.16 \\
                          &                                & SMMR Enhanced & \textbf{0.80} & \textbf{0.74} & \textbf{0.79} & \textbf{0.79} & \textbf{0.84} & \textbf{0.84} & 3.76 & 4.78 \\
\midrule
\multirow{4}{*}{Validation}& \multirow{2}{*}{GPT-3.5-turbo} & Baseline & 0.60 & 0.63 & 0.60 & 0.73 & 0.70 & 0.70 & 4.86 & 5.54 \\
                           &                                & SMMR Enhanced & \textbf{0.80} & \textbf{0.76} & \textbf{0.79} & \textbf{0.80} & \textbf{0.83} & \textbf{0.83} & \textbf{3.29} & \textbf{3.96} \\
                           & \multirow{2}{*}{GPT-4-turbo}   & Baseline & 0.83 & 0.75 & 0.81 & 0.81 & 0.81 & 0.81 & 2.54 & 3.08 \\
                           &                                & SMMR Enhanced & \textbf{0.83} & \textbf{0.79} & \textbf{0.82} & \textbf{0.82} & \textbf{0.85} & \textbf{0.85} & \textbf{2.43} & 3.37 \\
\bottomrule
\end{tabular}
\begin{tablenotes}
\footnotesize
\item Note: Acc, MF1, MPrec, and MRec stand for Accuracy, Macro F1, Macro Precision, and Macro Recall, respectively.
\end{tablenotes}
\end{threeparttable}
\end{table*}
\begin{table}[!t]
  \centering
  \begin{threeparttable}
    \caption{Mental Health Detection on Case Study Dataset}
    \label{tab:case_study_table}
    \fontsize{8}{11}\selectfont 
    \begin{tabular}{lccccc}
      \hline
      \textbf{Method} & \textbf{Valid (\%)} & \textbf{Acc.} & \textbf{F1} & \textbf{Ave.} & \textbf{SD} \\
      \hline
      GPT-4o Baseline      & 92   & 0.95 & 0.98 & 6.97 & 3.20 \\
      GPT-4o+SMMR  & \textbf{100}  & 0.93 & 0.97 & 6.85 & \textbf{2.90} \\
      GPT-3.5 Baseline      & 98   & 0.91 & 0.95 & 6.66 & 3.17 \\
      GPT-3.5+SMMR & \textbf{100}  & \textbf{0.93} & \textbf{0.97} & \textbf{7.03} & \textbf{2.57} \\
      GPT-4 Baseline        & 100  & 0.92 & 0.96 & 7.02 & 2.95 \\
      GPT-4+SMMR   & 100  & 0.91 & 0.95 & \textbf{7.40} & \textbf{2.67} \\
      \hline
    \end{tabular}
    \begin{tablenotes}
      \small
      \item Note: This table shows the results for both binary mental health detection (accuracy and F1 scores) and the correctness of disorder type identification (Ave, SD).
    \end{tablenotes}
  \end{threeparttable}
\end{table}

\section{Results}
\label{sec:results}

We evaluated the SMMR framework on both DAIC-WOZ and our curated case studies, comparing its performance against single-model baselines. Tables~\ref{tab:smmr} and \ref{tab:case_study_table} present the key metrics, reflecting a consistent advantage for SMMR-enhanced approaches.

\subsection{DAIC-WOZ Performance}
Table~\ref{tab:smmr} highlights the performance of GPT-3.5-turbo and GPT-4-turbo across training, testing, and validation splits. Alongside classification measures such as Accuracy (Acc.), F1, Macro F1 (MF1.), Macro Precision (MPrec.), Macro Recall (MRec.), and ROC AUC (for PHQ-8 $\geq 10$ classification), we also track Mean Absolute Error (MAE) and Root Mean Squared Error (RMSE) for PHQ-8 scoring. Overall, the SMMR-enhanced models outperform baselines on every split, with particularly notable improvements in binary classification accuracy and F1. For example, GPT-3.5-turbo sees its testing accuracy jump from 0.55 to 0.76, while MAE decreases significantly (6.04 to 4.22), suggesting better handling of nuanced long-context data.

Importantly, SMMR also boosts consistency across different subsets. The validation accuracy for GPT-3.5-turbo, for instance, jumps from 0.60 to 0.80, showing that leveraging multiple “experts” and iterative refinement reduces hallucinations and more accurately captures PHQ-8 severity signals.

\subsection{Case Study Dataset Performance}
Table~\ref{tab:case_study_table} summarizes results on our 48 narrative case studies, where we evaluate binary detection of mental health concerns and the correctness of disorder-type identification. Although the baseline metrics here are relatively high for GPT-4 variants (exceeding 0.90 accuracy), SMMR demonstrates small but meaningful gains and, in some configurations, yields a 100\% valid output rate. Notably, GPT-3.5 sees its accuracy rise from 0.91 to 0.93 under SMMR, while its F1 measure improves from 0.95 to 0.97. Beyond classification, SMMR also stabilizes disorder-type identification; for instance, the average correctness (Ave.) for GPT-4 climbs from 7.02 to 7.40, indicating richer extraction of relevant diagnostic details.

In summary, these findings confirm that layering multiple models curbs inconsistencies and incomplete responses, leading to more robust and thorough assessments in both interview transcripts and case narratives.

\section{Conclusion}
In this study, we introduced Stacked Multi-Model Reasoning (SMMR) as a layered solution for long-context mental health assessments. By progressively refining the outputs of multiple “expert” models, SMMR demonstrated notable improvements in diagnostic accuracy, F1-scores, and PHQ-8 estimation on both the DAIC-WOZ dataset and a curated set of case studies. These gains highlight how a structured, multi-expert approach can mitigate hallucinations, capture subtle clinical cues, and enhance consistency in sensitive, high-stakes domains.

Despite these encouraging results, several limitations warrant attention. First, our model diversity was limited to commercial GPT variants; although these models are powerful, our findings may not generalize to local or open-source LLMs without further experimentation. Second, mental health assessments impose strict confidentiality and ethical constraints. While SMMR can reduce errors and inconsistencies, thorough clinical validation and prospective studies remain necessary to confirm its real-world safety and efficacy. Additionally, the data scarcity inherent in mental health research restricts the breadth of our evaluation, and the computational overhead of running sequential layers can pose scalability challenges. Finally, conflict resolution among layered “experts” raises questions about decision transparency, bias, and the need for IRB-reviewed protocols when dealing with vulnerable populations.

Looking ahead, we plan to expand SMMR by refining conflict-resolution strategies, exploring multi-modal data (e.g., audio and video cues), and investigating cost-effective deployments that balance performance with resource usage. We also intend to collaborate with clinical professionals to ensure that SMMR’s layered outputs align with human judgment in ethically rigorous settings. Taken together, these developments could make SMMR a robust foundation for safer and more reliable AI-driven mental health screening, addressing the complexity and subjectivity that define this crucial application area.

\section*{Acknowledgment}
ChatGPT was used to revise the writing to improve the spelling, grammar, and overall readability.

\bibliographystyle{IEEEtran}
\bibliography{bibi0114}

\end{document}